# Exploring Text Virality in Social Networks

**Marco Guerini** and **Carlo Strapparava** and **Gözde Özbal**
FBK-Irst
Via Sommarive 18
Povo, I-38100 Trento
guerini,strappa,ozbal@fbk.eu

**Abstract**

This paper aims to shed some light on the concept of virality - especially in social networks - and to provide new insights on its structure. We argue that: (a) virality is a phenomenon strictly connected to the nature of the content being spread, rather than to the influencers who spread it (b) virality is a phenomenon with many facets, i.e. under this generic term several different effects of persuasive communication are comprised and they only partially overlap. To give ground to our claims, we provide initial experiments in a machine learning framework to show how various aspects of virality can be independently predicted according to content features.

## Introduction

Analyzing and recognizing the persuasive impact of communication is of paramount importance in many theoretical and applied contexts. For example, Persuasive NLP focuses on various techniques such as crowdsourcing (Guerini, Strapparava, and Stock 2010) and corpus based methods (Guerini, Strapparava, and Stock 2008) to address the various effects which persuasive communication can have in different contexts on different audiences.

In this scenario "virality" and data collected from social networks are burning topics for activities such as Viral Marketing - a set of marketing techniques using social networks to achieve various marketing objectives both in monitoring and promotion. *Buzz monitoring*, for example, is the marketing technique for keeping track of consumer responses to services and products. It involves checking and analyzing online sources such as forums, blogs, etc. in order to improve efficiency and reaction times. Moreover, identifying positive and negative customer experiences (*white* and *black buzz*), can help to assess product and service demand, tackle crisis management, foster reputation online, etc. On the other side, as a promotional strategic goal, Viral Marketing campaigns can have to sustain *discussion* to get visibility for their product or to provoke *controversial* reactions to the brand (e.g. to strengthen the membership feeling).

Generally speaking, virality refers to the tendency of a content to spread quickly in a community by word-of-mouth. In the spreading process there are several elements at play, e.g. the nature of the *spreader* and of the *audience*, the structure of the *network* through which the information is moving and the nature of the *content* itself.

In this paper, we will mainly focus on the last point. We will first argue that virality hinges on the nature of the viral content itself, and that it can be decomposed into several components. Then, we will provide some preliminary experiments, in a machine learning framework, using the Digg dataset described in (Paltoglou, Thelwall, and Buckley 2010). These experiments show how the various components of virality can be *separately* predicted using just the wording of the content being spread.

**Related Work.** Several researchers studied information flow, community building and similar tasks using Digg as a reference dataset (Lerman and Ghosh 2010; Jamali and Rangwala 2009; Khabiri, Hsu, and Caverlee 2009; Aaditeshwar Seth and Cohen 2008). However, the great majority considers only features related to the network itself or simple popularity metrics of stories (e.g. number of diggs, number of comments), without analyzing the correlations of these aspects with the actual content spreading within the network.

A notable work using Digg datasets that began to investigate some of the insights we proposed is (Jamali 2009). It incorporates features, derived from sentiment analysis of comments, to predict the popularity of stories. Finally, the work presented in (Berger and Milkman 2009) is probably the closest to the approach proposed in this paper. It uses *New York Times* articles to examine the relationship between the emotion evoked by content and virality, using semi-automated sentiment analysis to quantify the affectivity and emotionality of each article. Still, their virality metric only consists of how many people emailed the article, which is an interesting but very limited metric.

## Virality characteristics

Virality is a social phenomenon in which there are no "immune carriers"[1]. That is to say, a content is either viral or not. If a message is viral, it will immediately spread, by no chance it will remain latent, waiting to activate.

This means that although the study of epidemiology and information spreading in social networks is important, it does only partially account for the phenomenon of content virality. Individuating the central nodes (i.e. the influencers) of a social network is certainly necessary if we want our content to spread quickly to a significant audience, but it will not grant its spreading to further nodes. The book "The tipping point" (Gladwell 2002) maintains that influencers have a central role in the spreading of a fashion or a product - even to further nodes - and this seems to contradict our idea. However, Gladwell's standpoint has also been criticized by some researchers (Watts and Dodds 2007).

Thus, we can reformulate our standpoint in a less controversial way: Social network structure analysis accounts for *how* content spreads, rather than *why*. Accordingly, our first aim is to investigate the characteristics of the content which make it viral.

In this work we consider "virality" of text-based contents, focusing on the various effects created within an online community. Still, our analysis method can also be applied to other kinds of contents such as images and video-clips.

Traditional approaches based on "popularity metrics" (e.g. the number of *I_like*) are not sufficient for modeling a complex phenomenon such as "virality" of a text. First of all, let us try to distinguish different phenomena:

- *Virality*: in its basic form it refers to the number of people who accessed a given content in a given time interval. This is the most generic definition of virality.
- *Appreciation*: how much people like a given content, for example by clicking an *I_like* button.
- *Spreading*: how much people tend to share this content by forwarding it to other people.
- *Simple buzz*: how much people tend to comment a given content.
- *White buzz*: how much people tend to comment in a positive mood (e.g. "The best product I have ever bought").
- *Black buzz*: how much people tend to comment in a negative mood (e.g. "Do not buy this product, it is a rip-off").
- *Raising discussion*: the ability to induce discussion among users.
- *Controversiality*: the ability to split the audience in different parties (usually pro and against the given content).

While many of the above definitions are straightforward, some of them deserve a little more discussion (i.e. *black buzz*, *raising discussion*, *controversiality*).

A paradigmatic example of *black buzz* is the "Bonsai Kitten" hoax. In 2000 a website suggested that kittens were being subjected to horrific mistreatment to shape their bodies, as if trees were being shaped into bonsai. The "bonsai kitten" hoax was widely believed and provoked a storm of outrage and negative word of mouth reactions throughout the world, up to a point that FBI was compelled to investigate the Web site. Similarly, but less outrageously, the recent iPhone 4 antenna problem provoked a lot of talks on reception issues. For a while Apple ignored the issue, but the black buzz grew so quickly that Apple CEO, Steve Jobs, had to hold a press conference to address the problem promising a free of charge solution. These examples show how a content can spread exactly because it is disliked.

Some well know examples of *raising discussion* refer to movie promotion, namely "The Blair witch project" and "Cloverfield". Both movies managed to became a cultural phenomenon long before their release. Just consider that the discussion about the movie "The Blair witch project" rose a $30,000 film into a $150 million blockbuster. All these discussions were triggered by two very simple but powerful human characteristics: curiosity and mystery. People were discussing about its reality and trying to guess whether all the things shown in the movie had really happened.

As another example: "Medal Of Honor" is a video game set during the current Afghanistan war. The *controversy* raised about the option, for players, to play also as Taliban, splitting the audience in pro (claiming that it was only a game) and against (claiming that this was harmful and disrespecting for fallen Allied soldiers). The difference of opinions about the product was so strong that it echoed worldwide. More generally, there are topics which are intrinsically controversial such as death penalty and abortion.

## Dataset, Metrics and Experiments

In this section, we formalize some of the aforementioned phenomena using the Digg dataset described in (Paltoglou, Thelwall, and Buckley 2010). This corpus allows us to define and analyze many of the above definitions in a unique framework. These formalizations (metrics) represent an initial attempt to model phenomena which have not been addressed before, and they can be refined as a future work. Starting from these metrics we then build specialized datasets for our experiments.

**Digg Dataset.** Digg is a social bookmarking website where people can share, vote and comment "stories". Stories typically contain a `title`, a short textual description (`snippet`) and a link to the original web content. In this dataset the average length of the `title` is 7 words, the average length of `snippet` is 31 words. Users are allowed to publicly approve submitted stories with a process called "digging", and publicly approve or disapprove comments on submitted stories (called "DiggUp" and "DiggDown"). The data gathered from the site is a complete crawl spanning the months February, March and April 2009. Some statistics concerning the gathered data include: 1,195,808 submitted stories; 135,367 stories with at least one comments; 1,646,153 individual comments; 877,841 users who submitted a story/comment; 14,376,742 approvals (i.e. diggs).

**Appreciation.** Appreciation is connected to the number of

---
[1] We thank Francesco Varanini for the useful metaphor.

diggs a story received[2]. The formula for appreciation is:

$$A = ND \quad (A)$$

where $ND$ is the number of diggs for a given story.
*Dataset:* we considered as appreciated stories those receiving more than 100 diggs (i.e. $A \geq 100$). In the dataset there are 16660 stories with this characteristic. Then, we randomly extracted an equal number of unappreciated stories (i.e. 0 or 1 digg) to obtain a dataset of 33320 instances.

**Buzz and Spreading.** Within the Digg community, the same story can be submitted only once. Therefore it is difficult to model the concept of spreading. In our view, the number of *different* users commenting on a story is a good hybrid measure to model buzz and spreading. The formula for Buzz and Spreading is:

$$BS = NUC \quad (BS)$$

where $NUC$ is the number of such different users.
*Dataset:* we considered as buzzed only stories which have a $BS \geq 100$. There are 3270 examples of buzzed stories, and we randomly extracted an equal number of non buzzed stories (i.e. 0 comments) to obtain a dataset of 6540 instances.

**Raising-discussion.** The number of comments alone do not state whether there has been a discussion among users about a story; we need to distinguish low level comments (i.e. *explicit* replies to other comments) from top level ones. The formula for raising discussion is:

$$RD = (NC_L/NC_T) * NUC \quad (RD)$$

where $NC_T$ is the number of comments for a given story and $NC_L$ is the number of (low-level) comments, i.e. comments which are replies to other comments.
*Dataset:* we considered as positive examples only those stories which have a $RD$ score $> 50$. Accordingly, we collected 3786 examples of stories which raised discussion, and then we randomly extracted an equal number of negative examples to obtain a dataset of 7572 instances.

**Controversiality.** The problem of opinion controversiality has received little attention, a notable exception being (Carenini and Cheung 2008). While their model comprises multiple classes (i.e. votes in a -3 to +3 scale), our model is binary: only "I like" (+1) or "I don't like" (-1) votes are allowed. Since stories cannot receive negative votes in Digg, we decided to consider the "DiggsUp" and "DiggsDown" of the associated comments. If the highest number of positive votes among comments is A and the highest number of negative votes is B, the formula for controversiality is:

$$C = min(A, B)/max(A, B) \quad (C)$$

$min(A, B)$ denotes the smaller value of A and B, and $max(A, B)$ denotes the larger of the two. This measure ranges from a minimum of 0 (total agreement, everybody who voted either liked or disliked the comments) to a maximum of 1 (highest controversiality, the sample votes splitted exactly into two).

---

[2]A digg represents an acknowledgement of a story, regardless of being positive or negative (e.g. in the Bonsai Kitten example).

*Dataset:* we considered as controversial stories only those which have a controversial score (C) $\geq 0.9$. We collected 3315 examples of controversial stories, and we randomly extracted an equal number of non controversial stories to obtain a dataset of 6630 instances.

*Datasets preprocessing.* We focused only on the text present in the stories. As features we solely considered the words contained in the `title` and in the `snippet` of the story itself. To reduce data sparseness, we PoS-tagged all the collected words (using TreeTagger (Schmid 1994) and considered the lemmas only for content words, i.e. nouns, verbs, adjectives and adverbs). In the experiments, we did not make any frequency cutoff or feature selection.

*Experiments.* We conducted a series of preliminary experiments, in a machine learning framework, to explore the feasibility of independently predicting the impact of stories according to the various metrics proposed. For all the experiments we exploited Support Vector Machines framework, in particular we used SVM-light under its default settings (Joachims 1998). For every metric we conducted a binary classification experiment with ten-fold cross validation on the corresponding dataset. Note that all datasets are balanced, i.e. 50% of clearly positive examples and 50% of clearly negative examples - according to the metric under investigation: this accounts for a random baseline of 0.5.

|  | F1 measure |
|---|---|
| *Appreciation* | 0.78 |
| *Buzz* | 0.81 |
| *Controversiality* | 0.70 |
| *Raising-Discussions* | 0.68 |

Table 1: Results for the classification tasks

## Discussion

As far as the performance is concerned, we can state that the simpler metrics (namely *buzz* and *appreciation*) have much higher F1 (see Table 1). This can be explained by the fact that they represent simpler and more direct indicators of audience reaction to a given content. Still, also complex metrics (i.e. *raising-discussion* and *controversiality*) yield a good F1, which suggests that these phenomena can be automatically predicted as well, and that they are highly correlated to (and provoked by) the content being spread.

Next step is to understand to which extent the various viral phenomena are independent. If we consider Table 2 (where the percentage of the stories of a class overlapping with another is given) we can see that *buzz* and *appreciation* are the datasets with the greatest intersection (77% of buzzed stories are also "appreciated"). The correlation (and predictability) between the two phenomena has been in the focus of some research that confirms their interdependencies (Jamali and Rangwala 2009). Going into details of Table 2 we see that:

- Appreciated stories overlap quite similarly with buzz and raising-discussion stories (15.1% and 14.8%) while the controversial ones are the most independent (only 4.2%).

- For buzzed stories we see a high relationship with the appreciated ones (as stated), and a significant overlapping with raising-discussion ones (51,7%), but also in this case a very low overlapping with controversials (only 3.1%).
- For controversial stories we already saw a low relationship with appreciation and buzz (21.4% and 3.0% of the controversial stories) but a higher overlapping with raising-discussion ones (48.6%).
- Raising-discussion stories, instead, seem to overlap quite equally with the three other classes.

|  | App | Buzz | Cont | Rais |
|---|---|---|---|---|
| Appreciation | - | 15,1% | 4,2% | 14,8% |
| Buzz | 77,0% | - | 3,1% | 51,7% |
| Controversial | 21,4% | 3,0% | - | 48,6% |
| Raising | 65,0% | 44,6% | 42,5% | - |

Table 2: Percentage of stories overlapping with another class

**White and Black Buzz.** We tried to investigate the issues regarding white and black buzz. The Digg corpus contains, for each comment, a `BinaryEmotion` field, which yields the results of an automatic classifications using a Lexicon-based approach; see (Paltoglou et al. 2010) for further details. The `BinaryEmotion` field indicates whether the comment is considered as *positive*, *negative* or *neutral*. The formulae for White Buzz and for Black Buzz, using the `BinaryEmotion` field on the comment, are:

$$WB = (Positive > (Neutral + Negative)) \quad \text{(WB)}$$

$$BB = (Negative > (Neutral + Positive)) \quad \text{(BB)}$$

where *Negative*, *Neutral* and *Positive* indicate the number of comments for a given story, categorized as negative, neutral or positive, accordingly. We calculated the WB and BB for the stories with $NUC \geq 100$. According to the formulae proposed previously, we found 254 stories raising white buzz, and 1499 stories raising black buzz. These examples are too few for a machine learning approach and, furthermore, since the `BinaryEmotion` field is automatically calculated, results may not be very reliable. Nonetheless, it seems that a negative mood (black buzz) is predominant in the dataset. This predominance of black buzz can be accounted for by considering that the DIGG structure does not allow the expression of "diggsDown" for stories. So people could be more inclined to explicitly express (with comment) their dislike. Anyway this issue deserves further exploration.

## Conclusions

In this paper, we argued that content virality, in social networks, hinges on the nature of the viral content itself, rather than on the simple structure of the social network. We further argued that virality is a complex phenomenon which can be decomposed into several components. We defined and described many of these viral components, namely *appreciation*, *spreading*, *buzz*, *white* and *black* buzz, *raising discussion* and *controversiality*. We then provided experiments, in a machine learning framework, which showed how the various viral phenomena can be *separately* predicted using just the wording of the content being spread, and analysed the interdependencies amongst them.

As a future study we will deeply explore black and white buzz, possibly experimenting on other annotated datasets, and incorporating as features the emotions evoked by the stories (e.g. "fear", "joy", "anger"). Moreover, we are going to extract "viral" lexicons to be used in applicative scenarios, such as generating messages for marketing campaigns.